\pdfoutput=1
\documentclass[11pt]{article}
\usepackage{ACL2023}
\usepackage{times}
\usepackage{latexsym}
\usepackage[T1]{fontenc}
\usepackage[utf8]{inputenc}
\usepackage{microtype}
\usepackage{inconsolata}
\usepackage{pifont}
\newcommand{\cmark}{\ding{51}}%
\newcommand{\xmark}{\ding{55}}%

\newcommand{\ie}{\textit{i.e.\ }}

\usepackage{amsmath,amssymb} 
\usepackage{multirow}
\usepackage{graphicx}
\usepackage{booktabs}
\usepackage{subfigure}
\usepackage{enumitem}

\title{End-to-end Knowledge Retrieval with Multi-modal Queries}
\author{
    Man Luo$^1$ \quad Zhiyuan Fang$^2$ \quad Tejas Gokhale$^1$ \quad Yezhou Yang$^1$ \quad Chitta Baral$^1$ \\
    \textsuperscript{1} Arizona State University \quad \textsuperscript{2} Amazon Alexa\\
    \texttt{\{mluo26, tgokhale, yz.yang, chitta\}@asu.edu}, \quad \texttt{zyfang@amazon.com}
}

\begin{document}

\maketitle

\begin{abstract}
We investigate knowledge retrieval with multi-modal queries, \ie queries containing information split across image and text inputs, a challenging task that differs from previous work on cross-modal retrieval. 
We curate a new dataset called ReMuQ \footnote{pronounced \textit{re--$\mu$-queue}. Data and code: \url{https://github.com/luomancs/ReMuQ}.} for benchmarking progress on this task. ReMuQ requires a system to retrieve knowledge from a large corpus by integrating contents from both text and image queries. We introduce a retriever model ``ReViz'' that can directly process input text and images to retrieve relevant knowledge in an end-to-end fashion without being dependent on intermediate modules such as object detectors or caption generators. We introduce a new pretraining task that is effective for learning knowledge retrieval with multimodal queries and also improves performance on downstream tasks. We demonstrate superior performance in retrieval on two datasets (ReMuQ and OK-VQA) under zero-shot settings as well as further improvements when finetuned on these datasets. 
\end{abstract}

\section{Introduction}
Humans recall, retrieve, and communicate information using many indirect hints and cues. For instance, if we want to explain the concept of a ``leopard''  but have forgotten the name, we can relate the concept to a picture of a tiger and say ``it is an animal that looks like this, but has spots instead of stripes''.
Similarly, when children learn to draw a new shape like an \textit{oval}, teachers often prompt them by showing a circle, but saying ``make the circle stretched-out''.
This method of learning new concepts from visual aids and language descriptions is a common way of reinforcing existing knowledge and allowing learners to explore and retrieve new concepts \cite{kinder1942review}.

\begin{figure*}
    \centering
    \includegraphics[width=\linewidth]{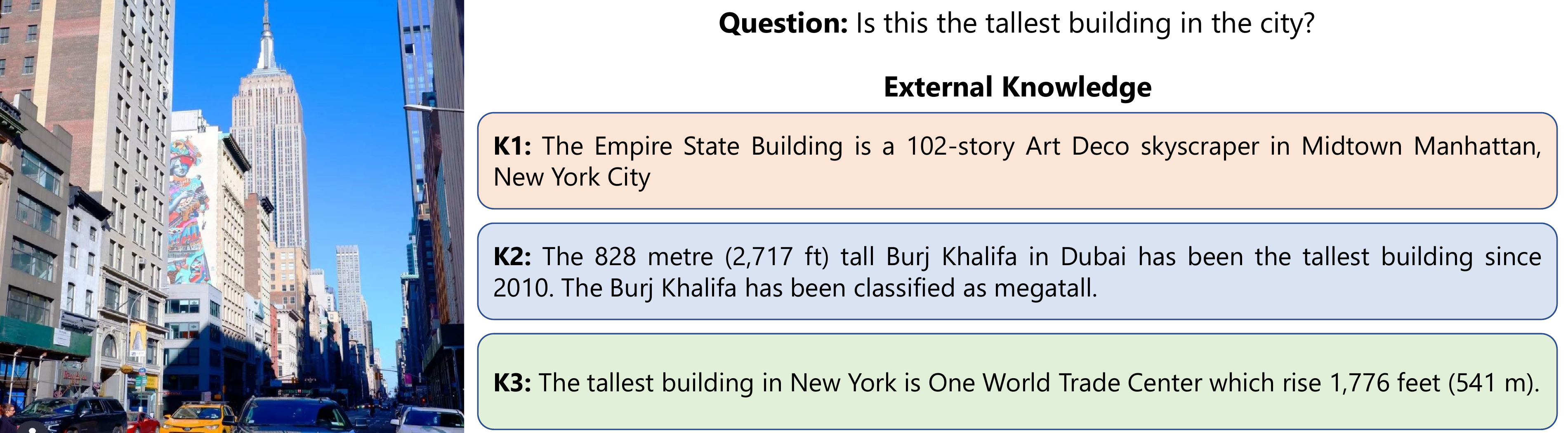}
    \caption{
    An illustration of the multimodal retrieval task from the ReMuQ dataset.  
    The image shows the Empire State Building and the question asks if it is the tallest building in ``the city''.
    Neither the image nor the question explicitly mentions that ``the city'' is New York.
    The challenge therefore is to use the cues in the image and question to retrieve relevant information and answer the question.
    In this illustration we show the retrieved knowledge using only the image (K1), only the question (K2), or both image and question (K3).
    Only K3 can be used to answer the question correctly.
    }
    \label{fig:task_teaser}
\end{figure*}

We propose a task for vision-language models to retrieve knowledge with multi-modal queries, i.e. queries in which hints about the information to be retrieved are split across image and text inputs.
Figure \ref{fig:task_teaser} contains an example of this task, where the image shows the Empire State Building in New York City.
If we retrieve knowledge using only the image, is it likely that the retrieved information (K1) will be related to the Empire State Building.
However, K1 is insufficient to answer the question.
On the other hand, if we retrieve knowledge using only the question, then the information retrieved (K2) is likely to be related to the tallest building in all cities (and not restricted to New York City).
K2 by itself is also insufficient to answer the question.
This example shows that the combined query containing both image and text (question) is necessary for retrieving relevant knowledge (K3).

We introduce a new benchmark and dataset called ReMuQ (\textbf{Re}trieval with \textbf{Mu}ltimodal \textbf{Q}ueries) to train and evaluate models to retrieve the answer from a corpus given multimodal (vision + language) queries.
To create multimodal queries, we start with the WebQA~\citep{chang2022webqa} dataset as a source -- WebQA contains images annotated with questions and answers.
We select questions from WebQA where the answer includes both an image and text.
We then remove any image information from text and combine the image and the augmented text to form a new multimodal query. 
We also construct a large retrieval corpus consisting of answer options of all questions as the source of knowledge for this task.

This task requires integrating the contents from both modalities and retrieve knowledge -- in this paper we denote such a system as a ``VL-Retriever''.
Existing  VL-Retrievers~\citep{qu2021passage,luo2021vrr,gao2022trig} typically follow a two-step process to retrieve knowledge: 
(1) converting the image into captions or keywords, appending them to the text query, and 
(2) using a text-retriever system to retrieve the knowledge. 
However, this approach can result in a loss of important information from the image, such as context and background.
Additionally, using a caption generation model trained on a particular domain does not transfer well to other domains in real-world applications.

To address these issues, we propose an end-to-end VL-Retriever that has the potential to leverage the entire image, rather than just object categories, keywords, and captions. 
We call this model \textit{ReViz}, a retriever model for  ``\textbf{Re}ading and \textbf{Viz}ualizing'' the query.
As part of ReViz, we use a vision transformer-based model, ViLT~\citep{kim2021vilt}, to directly encode the image from raw pixels with context inputs, and we employ BERT~\citep{devlin2018bert} as the knowledge encoder to represent the long, free-form text as a knowledge embedding.
ReViz differs from previous retrieval models in two main ways. First, it does not require an extra cross-modal translator (e.g., a captioning model) or object detector to represent the images. 
Second, its end-to-end design allows for the flexible retraining of each submodule of the model, which can mitigate potential issues caused by domain gaps.

Unlike neural text-retrievers~\citep{karpukhin2020dense,luo2022improving}, 
the query and knowledge encoders in ReViz are of different types of modality (\ie multimodal transformer and language transformer).
The different semantic spaces of the query and knowledge embeddings make alignment between them difficult. 
To address this, we propose a novel multimodal retrieval pretraining task.
To create training data, we construct triplets of (input-image, input-text, output-knowledge) from the WiT~\citep{srinivasan2021wit} dataset which contains encyclopedia-type knowledge from Wikipedia.
We process the data such that the input image and text have mutually exclusive information. 

\medskip
\noindent Our contributions and findings are listed below.
\begin{itemize}[noitemsep,leftmargin=*] 
    \item We introduce a new dataset \textit{ReMuQ} to facilitate research on retrieval with multimodal queries.
    \item We propose an end-to-end VL-Retriever, ReViz,  that directly acquires knowledge given multimodal query. ReViz is not dependent on any cross-modal translator, such as an image captioning model or an object detector.
    \item We pretrain ReViz on a novel multimodal retrieval pretraining task, VL-ICT. We observe that with the proposed pretraining on the WiT dataset, our VL-Retriever is a powerful zero-shot multimodal retriever that surpasses existing single-modal knowledge retrieval methods. 
\end{itemize}

\begin{figure*}[t]
    \centering
    \includegraphics[width=\linewidth]{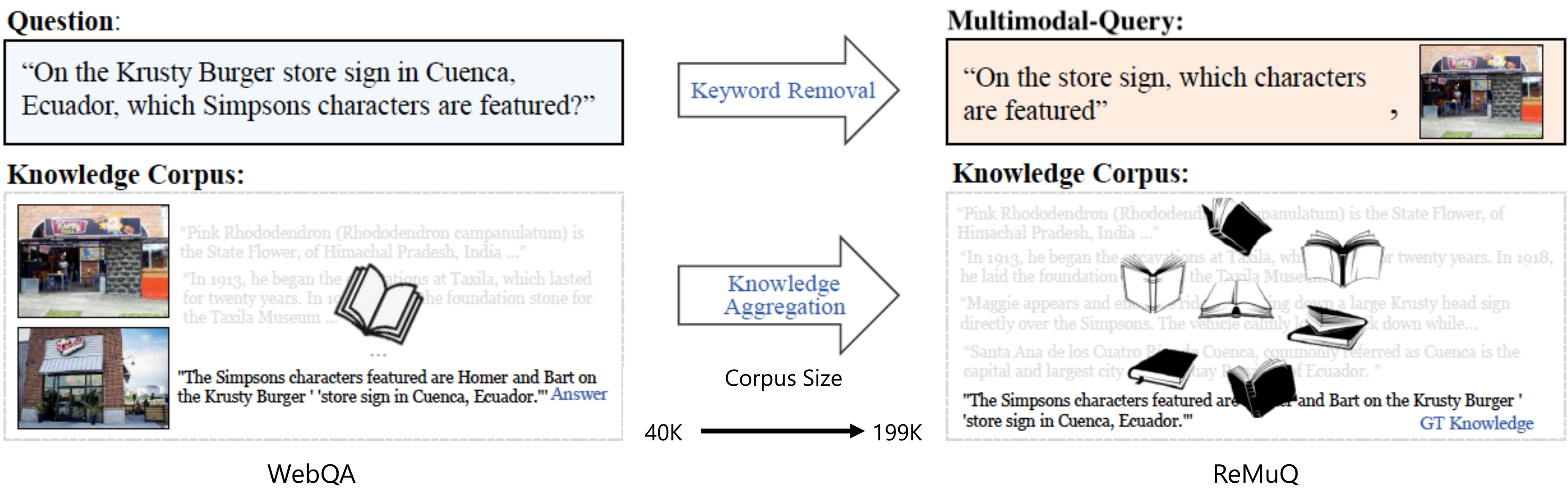}
    \caption{Dataset creation procedure for ReMuQ using WebQA as the source of the raw data. The multimodal-Query in ReMuQ is the combination of an image and the question from WebQA where the overlapped information with the image is removed. The ground truth knowledge of ReMuQ is the answer from WebQA. The corpus consists of all answers and the distracted knowledge candidates given in ReMuQ.}
    \label{fig:ReMuQ}
\end{figure*}

\section{Related Work}
\textbf{Cross-Modal Retrieval} aims to find information from a different modality than the query; for instance retrieving images from text (text-to-image), text from images (image-to-text) \cite{young2014image,lin2014microsoft}, text-to-video and video-to-text~\cite{rohrbach2015dataset,xu2016msr,zhou2018towards}. 
In contrast, we consider retrieval of knowledge for queries comprised of both modalities (i.e. image and text) together.

\medskip
\noindent\textbf{Knowledge-based Question Answering.}
Retrievers are important for finding relevant knowledge to aid knowledge-based question-answering models for tasks such as FVQA~\cite{wang2017fvqa} (commonsense knowledge), Text-KVQA~\cite{singh2019strings} which requires knowledge of the text in the image, and KVQA~\cite{shah2019kvqa}(world knowledge about named entities).
Both FVQA and KVQA are equipped with knowledge graph as external corpus.  
In OKVQA~\cite{marino2019okvqa} and its augmented versions S3VQA~\cite{jain2021select} and A-OKVQA~\cite{schwenk2022okvqa}, models are free to use any existing knowledge bases to retrieve relevant knowledge.
WebQA~\cite{chang2022webqa} is a multi-hop reasoning dataset that requires a system to aggregate multiple sources to answer a question, where the answers can be found either via image search or general web search.
\citet{fang2020video2commonsense} introduce a video question answering dataset that requires a system to answer questions using commonsense knowledge about intentions and effects of people's actions in videos.

\medskip
\noindent\textbf{Knowledge-Retrieval with Multimodal Queries}
While there are methods for retrieving knowledge from knowledge graphs~\citep{narasimhan2018out,li2020boosting,Marino2021krisp}, in this work, we focus on systems that retrieve knowledge from free-form text, which is more readily available and comprehensive. 
Previous methods involve converting images into language representations such as captions~\cite{qu2021passage,gao2022trig} or object tags~\cite{gui2021kat,yang2021empirical}, and then using a text-based retriever such as BM25~\cite{robertson2009bm25} or DPR~\cite{karpukhin2020dense} to find relevant knowledge. 
\citet{gao2022trig} leverage GPT-3~\cite{brown2020language} to generate the knowledge. 
\citet{qu2021passage,luo2021vrr} use a vision and language model to obtain cross-modal representations.
CLIP~\cite{radford2021clip} has also been applied to retrieval tasks; however it has limitations due to its separate encoding of text and image without a multi-modal fusion module.

\begin{figure*}
    \centering
    \includegraphics[width=\linewidth]{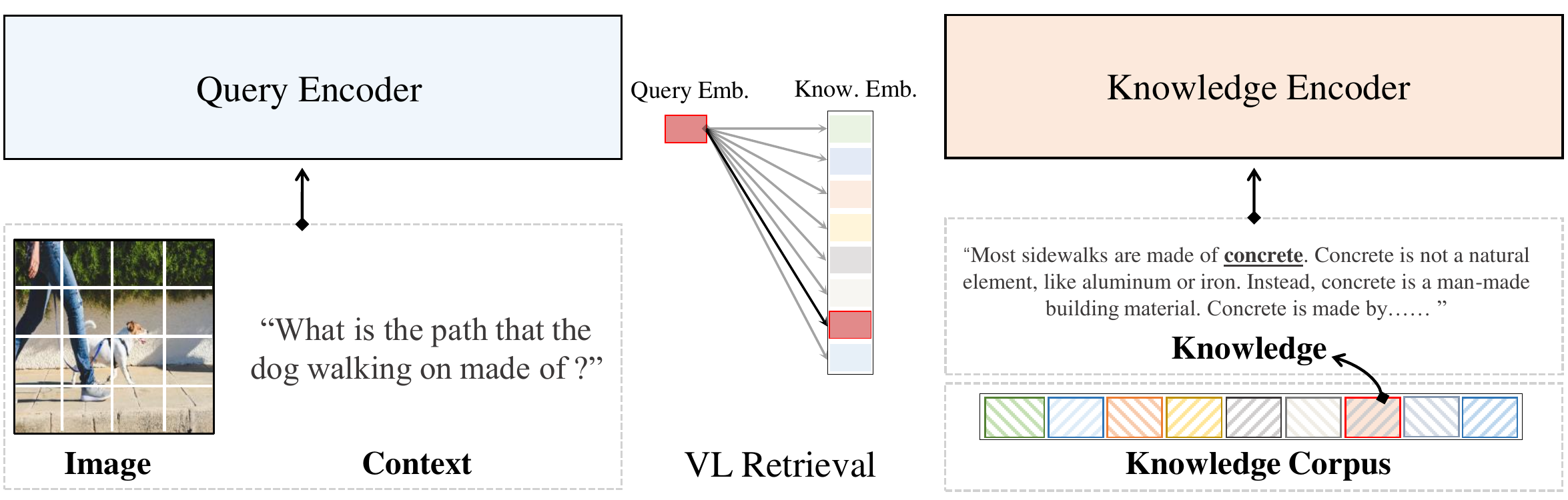}
    \caption{Overall architecture of ReViz. ReViz consists of a Vision-Language Transformer that encodes the image and text and a knowledge encoder that projects the knowledge into knowledge embedding. During inference, ReViz selects the knowledge from the corpus that has the largest relevance score with the image-text embedding.}
    \label{fig:arch}
\end{figure*}

\section{Retrieval with Multimodal Queries}
\label{sec:ReMuQ}
In this section, we define the problem statement for knowledge retrieval with multimodal queries and describe the construction of the ReMuQ dataset to assess models performing this task.

\subsection {Problem Statement} 
Given a query $Q = (I, T)$ containing as image $I$ and text $T$, we wish to learn a mapping to relevant textual knowledge $K$ from a corpus $C$.
Note that the two modalities $I$ and $T$ are such that each contains partial information about $K$.
Both $I$ and $T$ are necessary for successful retrieval of $K$ and Only using one of the two modalities is inadequate.

\subsection{ReMuQ Dataset Creation} 
In ReMuQ each query has exactly one ground truth knowledge associated with it.
To create such queries, we augment WebQA questions~\cite{chang2022webqa}, and collect a large corpus to serve as the knowledge source for any retrieval systems. 
WebQA is a multihop and multimodalQA dataset including text questions of different types such as Yes/No, Open-ended (e.g. shape, color, etc.), and multi choice (MC) questions.
The images are crawled from Wikimedia Commons, both  questions and text answers are created by annotators. 

To create multimodal queries, we utilize the MC questions in WebQA, which are associated with multiple choices as knowledge sources in the form of text or images. 
The ground truth answers of the questions include text-only, image-only, or both text and image.
We adapt important steps to create multimodal queries and explain the pipeline of the curation procedure below and in Figure~\ref{fig:ReMuQ} (more examples are given in Appendix). 

\medskip
\noindent\textbf{(1) Question Filtering.} 
    We select multiple-choice questions which have answer choices containing both image and text.

\medskip
\noindent\textbf{(2) Multimodal Query Construction.} The initial multimodal query is the combination of the question and the corresponding image. 
    In order to enforce a system to integrate information from both text and images, 
    we use \textit{tf-idf} to select keywords and then remove them in the question. 
    Our new multimodal-query is then the concatenation of the augmented question and the image, with the text-answer to be the ground-truth knowledge. 

\medskip
\noindent\textbf{(3) Retrieval Corpus Construction.}
    We aggregate the textual knowledge from all samples as the common knowledge corpus for multimodal retrieval, resulting in a large corpus of $\sim199k$ knowledge descriptions.

\medskip
\noindent\textbf{(4) Dataset Train-Test Split.}
    We divide ReMuQ into $70$\% for training and $30$\% as testing split.The new curated dataset contains $8418$ training samples and $3609$ testing samples, together with a knowledge corpus with $195,837$ knowledge descriptions. More statistic of ReMuQ is given in Table \ref{table:statistic}.

\begin{figure*}[t]
    \centering
    \includegraphics[width=\linewidth]{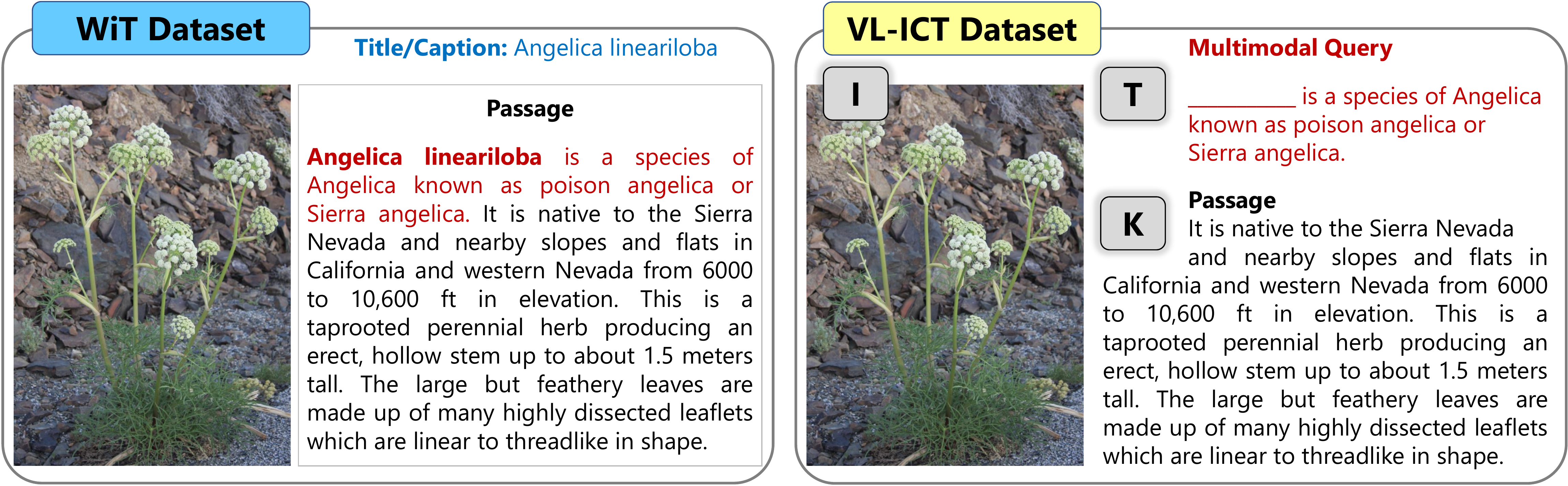}
    \caption{Figure on the left shows an example of the WIT dataset~\cite{srinivasan2021wit}, crawled from Wikipedia. 
    Figure on the right shows our constructed $({T,I,K})$ triplet: ${T}$ is a sentence from the passage and the words overlapped with the title/caption is masked; $K$ is the remaining passage after removing the sentence. }
    \label{fig:vl_ict}
\end{figure*}

\section{Method}
\label{sec:VLr}

Prior work on Vision-Language (VL)-Retrievers has focused on two-stage methods where the first stage involves feature-extraction using pretrained visual and textual encoders and the second stage learns retrieval using these features.
A typical VL-Retriever can be expressed as:
\begin{equation}
    K = \textsc{VL-Retriever}(T, F; C), 
\end{equation}
where $C$ is the knowledge corpus, $T$ is the text component of the query, and $F$ denotes the extracted features of image $I$.
This feature extraction can be done in two ways; (1) by converting the visual inputs into a human-readable textual description via an image captioning model or a series of object tags by object detector, (2) by extracting object features using an object detector.

\paragraph{End-to-End VL-Retriever.}
Instead, in this work, we are interested in building an end-to-end VL-Retriever, that encodes and selects the knowledge from the corpus using a VL model:
\begin{equation}
    K = \textsc{\textsc{VL}-\textsc{Retriever}}(T, I; C).
\end{equation}
We propose ReViz, an end-to-end \textsc{VL}-\textsc{Retriever} that learns to maximize the multimodal query and knowledge similarity for knowledge retrieval tasks. 
We introduce its architecture below.

\subsection{ReViz Model Architecture}
\label{sec:arch}
ReViz can read and visualize the input query, consists of two components, the multimodal query encoder and the knowledge encoder. 
Figure~\ref{fig:arch} illustrates the pipeline of our model. 

\paragraph{Multimodal Query Encoder.}
We use ViLT~\cite{kim2021vilt} to jointly encode the text input $T$ and the image $I$. 
In ViLT, an image is first partitioned into a set of a fixed size of patches -- these patches are encoded as continuous visual tokens through a linear projection layer~\cite{dosovitskiy2020image}.
These visual tokens are concatenated with the text tokens and summed with the position embeddings and fed into a stack of several self-attention blocks.
The final multimodal representation is obtained by applying linear projection and hyperbolic tangent upon the first index token embedding. 
\begin{equation}
    \mathbf{Z}_{q} = \text{ViLT} (I, T)
\end{equation}

\paragraph{Knowledge Encoder.}
To encode knowledge, we use a pre-trained BERT \citep{devlin2018bert} model, which produces a list of dense vectors $(h_1, \dots, h_n)$ for each input token, and the final representation is the vector representation of special token $\texttt{[CLS]}$.

\begin{equation}
    \mathbf{Z}_{k} = \text{BERT} (K)
\end{equation}
After the embeddings of query and knowledge are computed by the encoders, inner-dot product of the embeddings is considered as the relevancy score.
\begin{equation}
    \text{Score} (I, T, K)= \mathbf{Z}_{k}^{\top} \cdot  \mathbf{Z}_{q}
\end{equation}

\begin{table*}[t]
    \small
    \centering
    \begin{tabular}{lccccccc}
        \toprule
        \multirow{2}{*}{\textbf{Datasets}} &
        \multicolumn{2}{c}{\textbf{Source}} & 
        \multicolumn{2}{c}{\textbf{Average Length}} & 
        \multicolumn{3}{c}{\textbf{Size}} \\
        \cmidrule(lr){2-3} \cmidrule(lr){4-5}  \cmidrule(lr){6-8} 
        ~ & Image & Knowledge & Question & Knowledge & Train-D &  Test-D & Knowledge\\
        \midrule
        VL-ICT & Wiki & Wiki & 24.15 & 111.79 & 10,783,957 &  - &  - \\
        OKVQA & COCO & GS/Wiki & 9.15 & 67.05/100.00 & 8,958 &  5,046 & 112,724/21M \\
        ReMuQ & Wiki & Wiki & 14.97 & 48.60 & 8,418 & 3,609 & 195,837\\
        \bottomrule
    \end{tabular}
    \caption{A comparison of the datasets used in our experiment in terms of the sources of images and knowledge, average length of question and knowledge, and the sizes of each dataset. }
    \label{table:statistic}
\end{table*}

\subsection{Training}
\label{sec:train}
The training objective of ReViz draws inspiration from the instance discrimination principle based on contrastive learning.
The loss function to be minimized is given below:

\begin{equation}
    \label{eq:loss}
    \mathcal{L} = -\mathrm{log}~~
    \frac{\mathrm{exp}(\mathbf{z}_{q}\cdot\mathbf{z}_{k})}
    {
        \mathrm{exp}(\mathbf{z}_{q}\cdot\mathbf{z}_{k}) + \underset{\substack{\hat{\mathbf{k}}\in\mathbf{B}_{\mathbf{k}}, \hat{\mathbf{k}}\neq \mathbf{k}}}{\sum} \mathrm{exp}(\mathbf{z}_{q}\cdot \mathbf{\mathbf{z}_{\hat{k}}})}\ , 
\end{equation}
where $\mathbf{z}_{q}$ denotes the query embedding, $\mathbf{z}_{k}$ denotes the relevant knowledge embedding, and $\mathbf{z}_{\hat{k}}$ is the irrelevant knowledge embedding which serves as negative instances. 
We use all in-batch samples ($\mathbf{B}_{\mathbf{k}}$) as the negative instances.

\paragraph{Training with Hard Negatives.}
Adopting random samples as negative instances may cause sub-optimal metric space. 
Existing work shows that mining with hard negative samples leads to discriminative representations and has been applied to a broad series of tasks like face recognition~\cite{zhang2017range}, object detector~\cite{shrivastava2016training}, and metric learning for retrieval tasks~\cite{faghri2017vse++,harwood2017smart}.
Inspired by this, we also experiment with the hard negative technique to further boost the retrieval performance. 
To obtain the meaningful hard negative samples, we first train ReViz with the supervisions in \textit{eq.}~\ref{eq:loss}. 
With that, for each training question, we retrieve the top-$100$ knowledge instances (excluding the ground-truth) as the hard negative samples.
Note that we only apply hard negative mining to fine-tuning on downstream task but not the pretraining task (introduced in the next section).

\section{Pretraining Task for VL Retriever}\label{sec:vl-itc}
\label{sec:vlic}
Previous work~\citep{Chang2020PretrainingTF,lee2019latent,guu2020retrieval} suggests that pretraining a retriever on unsupervised task that closely resembles retrieval can greatly improve the downstream tasks performance.
We propose a pretraining task called VL-ICT, which is inspired by ICT~\cite{lee2019latent} task in NLP domain.  

\medskip
\noindent\textbf{ICT} aims to train text-based information retrieval (IR) system for the open-domain question answering task. 
To train a model without annotated data, \citet{lee2019latent} propose to construct pseudo $(question, context)$ pairs as the training data for IR system. 
In particular, given a passage ${P}$, a random sentence ${S}$ in the passage is selected as the pseudo question, and the remaining passage ${P'}$ is considered as the relevant context. 
Such a weakly-supervised setting enables large-scale ICT pre-training, leveraging any available knowledge base as the training corpus.

\paragraph{VL-ICT.} 
We propose VL-ICT task to pre-train ReViz, which can be applied to multi-modal scenarios when both language and vision inputs exist in the query.
In VL-ICT, a {$(I, T, K)$} triplet is used for training.
Importantly, $I$ and $T$, contain mutually exclusive information and are both necessary for knowledge retrieval. However, such condition is not naturally existing, thus, we propose an automatic procedure to construct triplet satisfying this condition in the following. 

\paragraph{VL-ICT Training Data.}
Figure~\ref{fig:vl_ict} shows a snapshot of our data construction process where we use the WiT dataset~\cite{srinivasan2021wit} as the source.
Each WiT entry provides a title of the page or an image caption, a passage, and an image. 
We use the image from this WiT entry as the image $I$ in our VL-ICT triplet. 
We observe that the title or caption is usually entities, it allows us to simply use word matching to find the sentences in the page passage that include the title/caption. 
We take such sentences as the text (${T}$), then we remove this sentence from the passage and use the remaining passage as the knowledge (${K}$).
To enforce that (${T}$) and  (${I}$) have mutually exclusive but important information, we mask keywords in ${T}$ that appear in both $T$ as well as the caption.
In our experiments, we only select the English entities in WiT and execute the above process, and this results in $3.2$ million ($I, T, K$) training triplets.

\section{Experiments and Results}
\begin{table*}
\centering
\small
\begin{tabular}{lccccccccc}
\toprule
\multirow{2}{*}{\textbf{Model}} & \multirow{2}{*}{\textbf{Dataset}} & \multirow{2}{*}{\textbf{KB-Size}} & \multicolumn{7}{c}{\textbf{Metric}} \\
\cmidrule(lr){4-10} 
~ & ~ & ~  & MRR@$5$  & P@$5$ & R@$5$ & R@$10$ & R@$20$ & R@$50$ & R@$100$  \\
\midrule
CLIP-IMG+Q  & OKVQA & GS-$112$K & $19.08$ & $11.13$ & $34.54$ & $50.48$ & $65.08$ & $80.62$ & $88.11$  \\
BM25 (GenCap) & OKVQA & GS-$112$K & $36.36$ & $27.54$ & $51.35$ & $63.04$ & $73.37$ & $84.21$ & $90.39$\\
DPR (GenCap) & OKVQA & GS-$112$K & $39.15$ & $27.72$ & $55.56$ & $66.44$ & $75.59$ & $87.17$ & $92.42$ \\
ReViz+VL-ICT & OKVQA  & GS-$112$K & \boldmath{$45.77$} & \boldmath{$33.18$} & \boldmath{$64.05$} &\boldmath{$75.39$} & \boldmath{$84.21$} & \boldmath{$91.64$} & \boldmath{$94.59$} \\
\midrule
TRiG~\cite{gao2022trig} & OKVQA & Wiki-$21$M & - & - &$45.83$ & $57.88$ & $72.11$ & $80.49$ & $86.56$  \\
CLIP-IMG+Q & OKVQA & Wiki-$21$M & $16.45$ & $9.66$ & $29.81$ & $43.00$ & $55.73$ & $72.73$ & $82.26$ \\
BM25 (GenCap)& OKVQA  & Wiki-$21$M & $36.43$ & $27.89$ & $50.16$ & $60.92$ & $71.62$ & $82.82$ & $88.74$\\
DPR (GenCap) & OKVQA & Wiki-$21$M & $41.15$ & $28.10$ & $59.41$ & $71.13$ & $81.73$ & $89.90$ & $93.39$\\
ReViz+VL-ICT & OKVQA & Wiki-$21$M & \boldmath{$44.03$} & \boldmath{$32.94$} & \boldmath{$62.43$} & \boldmath{$73.44$ }& \boldmath{$82.28$} & \boldmath{$89.93$} &\boldmath{$93.76$}\\
\midrule
CLIP-IMG+Q  & ReMuQ & $199$K & $0.34$ & $0.17$ & $0.78$ & $1.36$ & $2.41$ & $7.34$ & $47.88$  \\
BM25 (GenCap) & ReMuQ & $199$K  & $3.80$  & $5.59$  & $8.78$ & $10.75$ & $12.88$ & $15.88$ & $17.98$ \\
DPR (GenCap) & ReMuQ & $199$K & \boldmath{$31.23$} & \boldmath{$35.79$} & \boldmath{$43.42$} & \boldmath{$48.77$} &\boldmath{ $54.47$} & $61.40$ & $67.30$\\
ReViz+VL-ICT & ReMuQ  & $199$K  & $23.61$ & $29.52$ & $39.43$ & $46.77$ & $53.56$ &\boldmath{ $63.70$} & \boldmath{$71.13$}\\
\bottomrule
\end{tabular}
\caption{Zero-shot performance of ReViz and baselines on two datasets: OKVQA and ReMuQ. OKVQA is evaluated on two knowledge sources. ReViz shows superior zero-shot performance in majority of the cases.}
\label{table:zeroshot}
\end{table*}
\begin{table*}[t]
    \centering
    \small
    \begin{tabular}{lccccccccc}
    \toprule
    \multirow{2}{*}{\textbf{Model}} & \multirow{2}{*}{\textbf{Dataset}} & \multirow{2}{*}{\textbf{KB-Size}} & \multicolumn{7}{c}{\textbf{Metric}} \\
    \cmidrule(lr){4-10} 
    ~ & ~ & ~  & MRR@$5$  & P@$5$ & R@$5$ & R@$10$ & R@$20$ & R@$50$ & R@$100$  \\
    \midrule
    ReViz & OKVQA & GS-$112$K   & $46.92$ & $34.51$ & $66.05$ & $77.80$ & $86.33$ & $93.34$ & $95.90$ \\
    ReViz+VL-ICT & OKVQA & GS-$112$K & \boldmath{$54.47$ }& \boldmath{$41.74$} & \boldmath{$73.35$} & \boldmath{$83.17$} &\boldmath{$89.56$} & \boldmath{$94.73$} & \boldmath{$96.81$}\\
    \midrule
    ReViz & OKVQA & Wiki-$21$M  & $41.66$ & $30.08$ & $60.88$ & $72.20$ & $81.07$ & $89.16$ & $93.10$  \\
    ReViz+VL-ICT & OKVQA & Wiki-$21$M  & \boldmath{$43.68$} & \boldmath{$31.36$} & \boldmath{$61.91$} & \boldmath{$72.63$} & \boldmath{$81.05$} & \boldmath{$89.28$} & \boldmath{$93.44$}\\
    \midrule
    ReViz & ReMuQ & 199$K$ & $41.03$ &  $49.08$ & $62.40$ & $71.63$ & $78.92$ & $86.60$ & $92.17$ \\
    ReViz+VL-ICT & ReMuQ & $199$K &\boldmath{$53.39$}& \boldmath{$62.11$} & \boldmath{$76.23$} & \boldmath{$83.32$} & \boldmath{$88.56$} & \boldmath{$93.41$} & \boldmath{$96.12$} \\
    \bottomrule
    \end{tabular}
    \caption{Comparison of ReViz when it is fine-tuned on downstream tasks. We compare ReViz and ReViz+VL-ICT (our pretraining task). VL-ICT enables ReViz to be a stronger multimodal-query retrieval model.}
    \label{table:reviz_ftune}
\end{table*}
\begin{table*}[t]
\small
\centering
\setlength\tabcolsep{5pt}
\begin{tabular}{lccccccccc}
\toprule
\multirow{2}{*}{\textbf{Model}} & \multirow{2}{*}{\textbf{FT}} & \multirow{2}{*}{\textbf{KB-Size}} & \multicolumn{7}{c}{\textbf{Metric}} \\
\cmidrule(lr){4-10} 
~ & ~ & ~ & MRR@$5$  & P@$5$ & R@$5$ & R@$10$ & R@$20$ & R@$50$ & R@$100$  \\
\midrule
VRR-IMG~\cite{luo2021vrr} &\cmark & GS-$112$K  & - & $31.80$ & $62.52$ & $73.96$ & $83.04$ & $90.84$ & $94.67$ \\
VRR-CAP~\cite{luo2021vrr} &\cmark & GS-$112$K  & - & $39.42$ & $71.52$ & $81.51$ & $88.57$ & $94.13$ & ${96.95}$ \\
ReViz+VL-ICT &\cmark & GS-$112$K & \textbf{54.47 }& \textbf{41.74} & \textbf{73.35} & \textbf{83.17} &\textbf{89.56} & \textbf{94.73} & $96.81$\\
\midrule
TRiG~\cite{gao2022trig} &\xmark & Wiki-$21$M & - & - &$45.83$ & $57.88$ & $72.11$ & $ 80.49$ & $86.56$  \\
ReViz+VL-ICT &\xmark & Wiki-$21$M  & \textbf{44.03} & \textbf{32.94} & \textbf{62.43} & \textbf{73.44 }& \textbf{82.28} & \textbf{89.93} &\textbf{93.76}\\
\bottomrule
\end{tabular}
\caption{Comparison of our best model with existing models on OKVQA. ``FT'' denotes fine-tuning. Our model surpasses existing methods by significant margins with or without fine-tuning and with different knowledge corpus.}
\label{table:okvqa_result}
\end{table*}

\paragraph{Datasets.}
In addition to ReMuQ, we conduct experiments on OKVQA to obtain stronger evidence for the efficacy of our method. 
Here, instead of QA task, we use OKVQA as a testbed for retrieval task, i.e. to retrieve a relevant knowledge to a question such that it contains the answer span. 
Furthermore, we use two corpora, a small corpus collected from Google search API introduced in ~\citet{luo2021vrr}, and a large corpus which contains 21M Wikipedia knowledge used in \citet{gao2022trig}. 
The statistic of each dataset is given in Table \ref{table:statistic}.

\paragraph{Evaluation Metrics.}
Following \citet{gao2022trig, luo2021vrr}, 
we evaluate the performances of models by Precision@K (P@K), Recall@K (R@K), and MRR@5.
We use similar metrics to evaluate the ReMuQ challenge except that P@$1$ is used instead of P@$5$ since ReMuQ has exactly one correct knowledge per query.

\begin{figure*}[t]
    \centering
    \includegraphics[width=\linewidth]{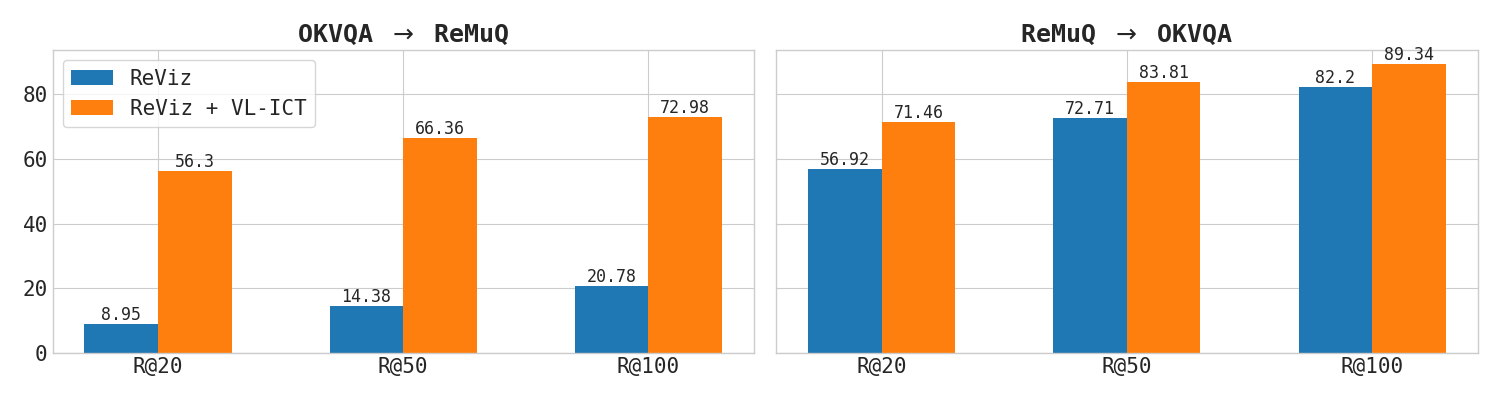}
    \caption{Evaluation of out-of-domain performances of ReViz and ReViz+VL-ICT. For OKVQA, we retrieve knowledge from GS-112K corpus. VL-ICT substantially improves the generalization of ReViz. Other metrics are given in Appendix. X->Y denotes using X as the training domain and Y as the testing domain.}
    \label{table:cross_domain_result}
\end{figure*}
\begin{figure*}
    \centering
    \includegraphics[width=\linewidth]{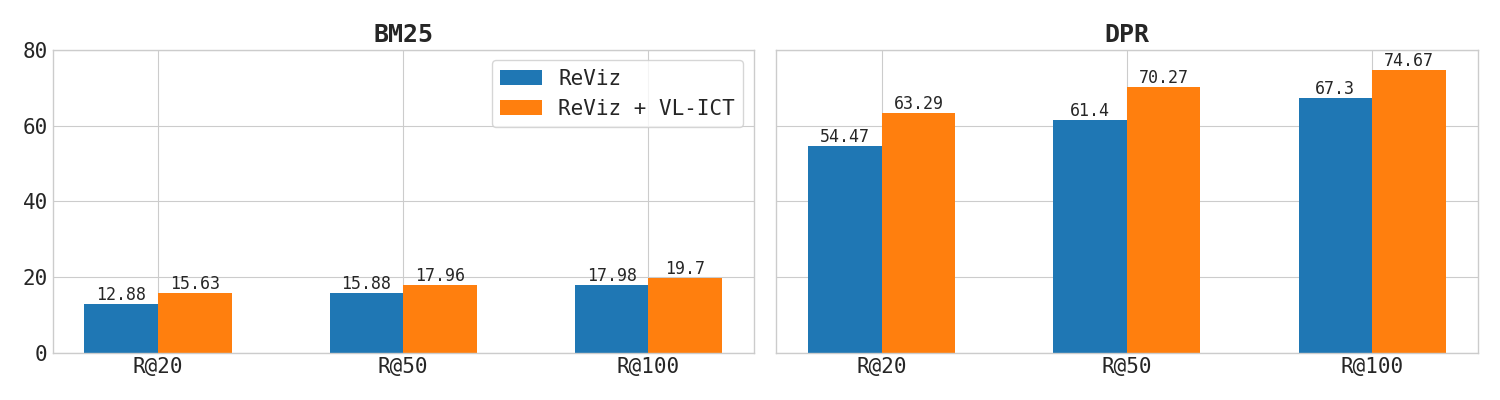}
    \caption{Comparison of captioning-dependent retrievers using generated captions and ground truth captions. The ground truth captions always lead to better performance than generated caption.}
    \label{table:remuq_cap}
\end{figure*}
\begin{figure}[t]
    \centering
    \includegraphics[width=\linewidth]{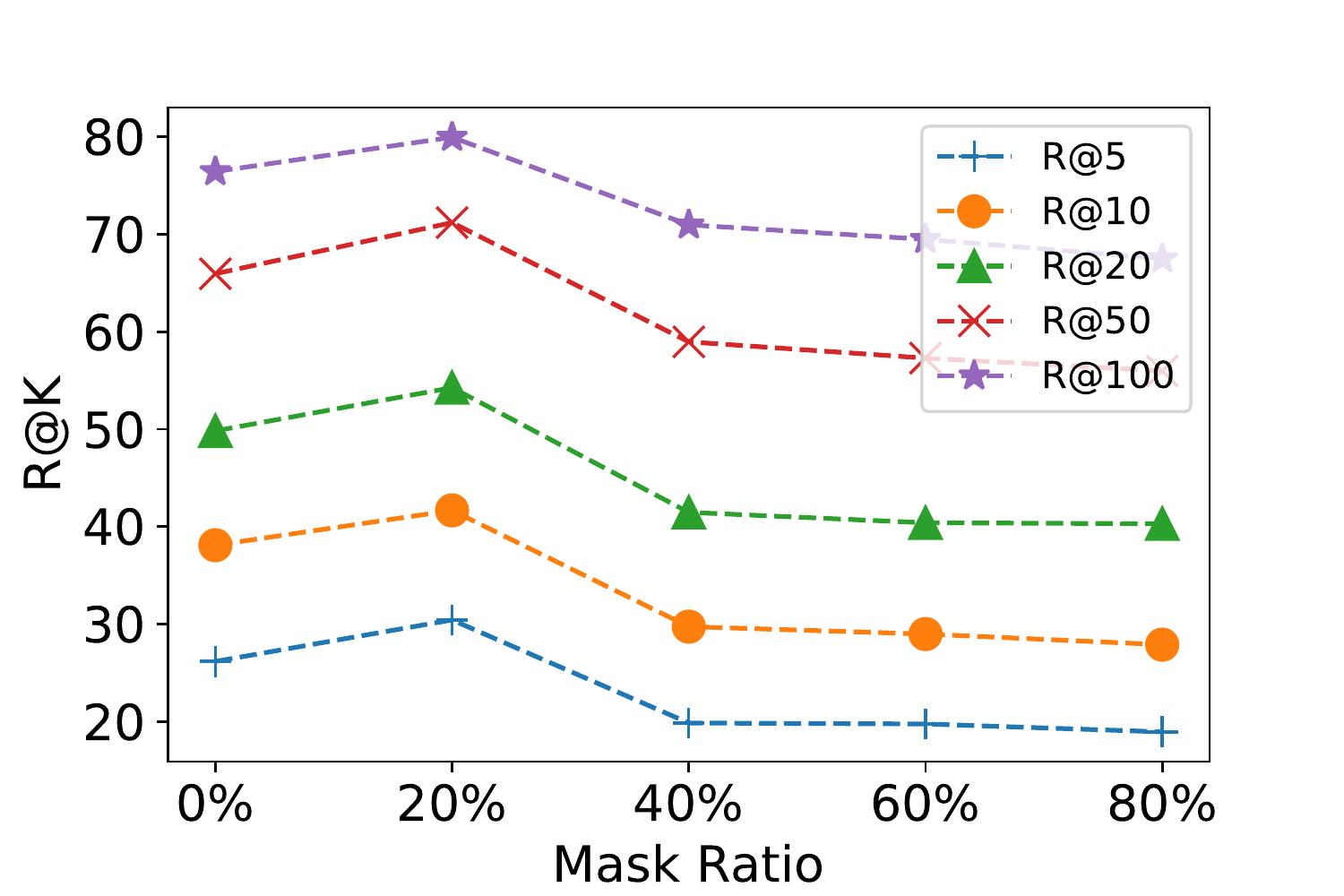}
    \includegraphics[width=\linewidth]{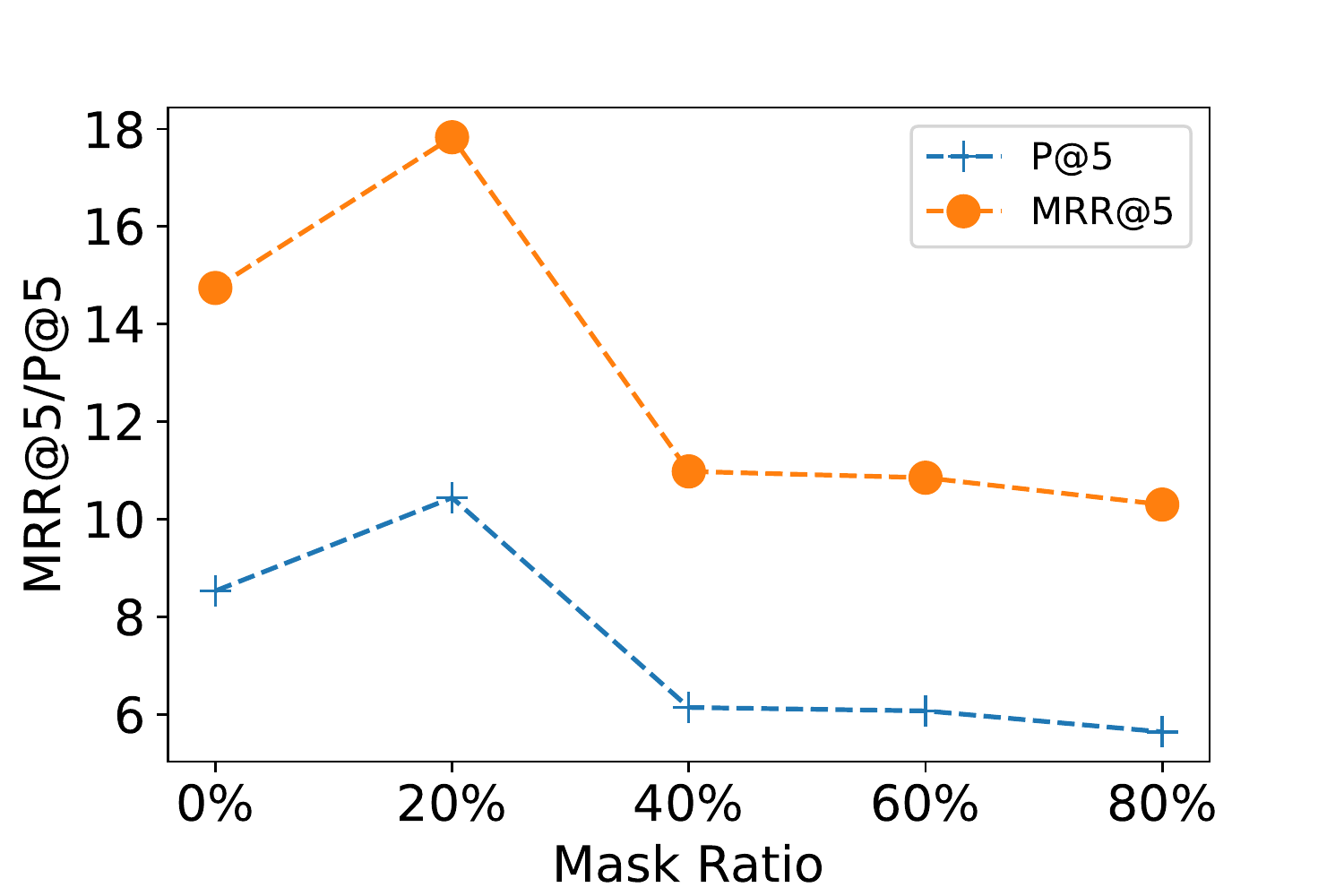}
    \caption{Effect of the masking ratio of sentences in VL-ICT task on ReViz's performance on OKVQA Task. We use GS112K as the knowledge corpus.}
    \label{fig:mask_ratio_recall}
\end{figure}
\subsection{Zero-shot Retrieval}
We first introduce three zero-shot baselines and then present the results.

\paragraph{CLIP Baseline.}
CLIP~\cite{radford2021clip} is a vision-language model pre-trained on over $400$M image-text pairs. 
We encode all knowledge descriptions via CLIP's textual encoder $\mathbf{K}$. 
Then, given an image-text pair as the query, we use the image encoder to get the visual representations ($\mathbf{I}$) and use the textual encoder to get the embedding of $\mathbf{Q}$. 
We compute the inner-dot products between all encoded visual representations ($\mathbf{I}$) and $\mathbf{K}$ to get the top-$100$ knowledge for evaluation, similarly for $\mathbf{Q}$. 
Finally we sum the scores and re-rank the top-100 knowledge. 
We find this performs the best than using individual modality (see Appendix). 

\paragraph{BM25 Baseline.} BM25~\cite{robertson2009bm25} is a well-known efficient retrieval algorithm for text-based retrieval task based on the sparse representation. We use the caption of the image to represent the information of the image and thus we convert the multi-modal knowledge retrieval task into a pure text-based retrieval task.

\paragraph{DPR Baseline.} 
We adopt DPR~\cite{karpukhin2020dense} trained on NaturalQuestions~\cite{kwiatkowski2019nq} dataset as a baseline, to retrieve the knowledge given an input image-text pair.
First, we use the contextual encoder of DPR to index the corpus, then we concatenate the question and the caption of the image as a joint textual query. 
With that, the question encoder of the DPR extracts the dense representation of the query for later computation. Lastly, we retain the most relevant knowledge pieces by calculating the inner-dot product between the query and the knowledge embedding.

\paragraph{Results.}
Table~\ref{table:zeroshot} shows the performances of baselines as well as ReViz pretrained on VL-ICT task. 
Among the baselines, we see that DPR is the strongest baselines. 
Surprisingly, although CLIP has shown strong performance on many classification and cross-modality pretraining task, it does not perform well on multimodal query retrieval task, this suggests that multimodal query retrieval is a challenging task for VL model. 
More importantly, we observe clearly that ReViz outperforms the baselines in terms of all metrics on OKVQA task on corpus of small and large size. On the ReMuQ dataset, ReViz wins CLIP and BM25 on all metrics, and DPR on two metrics.   
This demonstrates the effectiveness of our proposed pretraining task and the model design.

\subsection{Fine-tuning on Downstream Tasks}
To further demonstrate the effectiveness of VL-ICT pretraining task, we finetune models on downstream tasks and compare performance. We compare two versions of ReViz: (1) ReViz directly trained on the downstream task and (2) ReViz first pretrained on VL-ICT and then finetuned the down-stream task.
In addition, We study two senarios: in-domain, where a model is trained on the training set of X domain and evaluated on the testing set of X; out-of-domain, where a model is trained on the training set of X domain and evaluated on the testing set of Y domain.  

\paragraph{In-Domain Results.}
Table \ref{table:reviz_ftune} shows the in-domain performance. 
On both datasets, pretrained ReViz consistently outperform vanilla ReViz, suggesting that the pretraining task equips ReViz better alignment between the multimodal queries and the relevant knowledge.

\paragraph{Out-of-Domain Results.}
We investigate if the VL-ICT pretraining task can improve the generalization of ReViz.
We study the performances of ReViz under two settings: train on OKVQA (domain \textbf{X}) and test on ReMuQ (domain \textbf{Y}); and the inverse. 
Table~\ref{table:cross_domain_result} shows that ReViz+VL-ICT+\textbf{X} shows obviously better results than ReViz+\textbf{X} on \textbf{Y}, especially when \textbf{X} is OKVQA and \textbf{Y} is ReMuQ. 
This suggests that models pre-trained with VL-ICT tasks are more robust than models without VL-ICT.
We also see that the generalization performance still has a large gap with the fine-tuning, which suggests that OKVQA and ReMuQ are quite different tasks, and ReMuQ can be a good complement to OKVQA to study multimodal query retrieval task. 

\subsection{Comparison with Existing Methods}
We compare ReViz with existing retrieval methods for the OKVQA task. 
Note that most of the models on the leaderboard of OKVQA only report the final question answering accuracy but not the retrieval performance. 
In our experiments we include systems which report the retrieval performance. 

\paragraph{Baselines.}
\citet{luo2021vrr} present two fine-tuned multimodal retrievers:
VRR-IMG which uses LXMERT~\cite{tan2019lxmert} and VRR-CAP to convert the image into captions for knowledge retrieval. 
Both retrievers use GS-112K as the knowledge corpus. 
TriG~\cite{gao2022trig} uses zeroshot retriever and Wikipedia 21M as the knowledge corpus. 
Since these systems use either fine-tuned retriever or zero-shot retrievers, for fair comparison, we compare the best fine-tuned model and zeroshot model with the corresponding corpus.

\paragraph{Results.}
In the fine-tuning scenario, in majority of the cases (only one exception, R@100), our models consistently shows better performance than previous methods overall metrics.
Similarly, in the zero-shot case, our model is better than previous model on all metrics by large margins.

\subsection{Effects of Mask Ratio in VL-ICT Task}
In VL-ICT, we mask the keywords in the sentence to prevent information leakage.
Despite this, we find that the certain masked sentences still somehow overlap with the retrieved knowledge.
We conjecture that this overlapping makes the VL-ICL task inevitably easy, and thus impairs the effects of pre-training.
To study the optimal mask ratio, we conduct experiments to randomly mask the words in the sentence by different ratios. 
This study is performed on a smaller corpus of $1$ million VL-ICT training triplets and models are trained for one epoch.
Figure~\ref{fig:mask_ratio_recall} shows the results.
We observe that removing $20$\% of the keywords yields the best performance amongst all ratios and is also better than maintaining the sentences intact (0\% masking).

\subsection{Effect of Generated Captions}
Previous systems which rely on the caption generation model are affected by the quality of generated captions.
This may hamper the retrieval performance when the caption generation model is not trained on the same domain as the downstream task.
In our ReMuQ dataset, the images are from Wikipedia, but the caption generator is trained on MS-COCO~\citep{lin2014microsoft}.
We compare our two baselines, BM25 and DPR, using ground-truth image captions and the generated captions.
Table \ref{table:remuq_cap} shows that using the ground truth caption is much better than the generated caption in all cases. 
This suggests that the caption generator is the bottleneck of the retrieval methods to convert the image information to image captioning.
This demonstrates the limitations of previous methods and justifies our exploration of end-to-end training.

\section{Conclusion}
We study knowledge retrieval with multimodal (vision and language) queries, which, compared with existing retrieval tasks, is more challenging and under-explored. 
In addition, multimodal-query information retrieval has numerous potential applications, not only in retrieval tasks such as image, text, and video retrieval, but also in question answering, recommendation systems, and personal assistant.
The proposed dataset (ReMuQ) is ideally positioned to support the development of such functionalities.
We propose an end-to-end VL-retriever model, ReViz, which does not rely on any intermediate image to text translation modules.
A novel weakly-supervised task (VL-ICT) is proposed to enable large-scale pre-training.
Extensive evaluations on ReMuQ and OK-VQA datasets demonstrate that ReViz exhibits strong performance amongst all retrieval models in both zero-shot and fine-tuning scenarios.
Our proposed dataset and model provide a foundation for future work which could potentially lead to new findings and innovative applications in multimodal-query information retrieval.

\clearpage

\section*{Limitations}
During the creation of the ReMuQ dataset, we simply remove the words in the question that are duplicated in the image caption -- in some cases, this may result in grammatical errors in the text query. 
We performed the experiments for studying optimal masking ratio on a subset of the pretraining data, due to resource constraints.

\section*{Acknowledgments}
This work was supported by grants from National Science Foundation \#1816039 and \#2132724 and DARPA W911NF2020006.
The views and opinions of the authors expressed herein do not necessarily state or reflect those of the funding agencies and employers.
\bibliography{man}
\bibliographystyle{acl_natbib}

\section*{Appendix}
\appendix

\begin{table*}[t]
    \small
    \centering
    \resizebox{0.75\linewidth}{!}{
    \begin{tabular}{lcccccccc}
        \toprule
        \multirow{2}{*}{\textbf{Model}} & \multirow{2}{*}{\textbf{Dataset}} & \multicolumn{7}{c}{\textbf{Metric}} \\
        \cmidrule(lr){3-9} 
        ~ & ~  & MRR@$5$  & P@$K$ & R@$5$ & R@$10$ & R@$20$ & R@$50$ & R@$100$  \\
        \midrule
        CLIP-IMG & OKVQA & $18.96$ & $11.03$ & $34.50$ & $50.48$ & $65.12$ & $80.60$ & $88.11$ \\
        CLIP-Q & OKVQA  & 5.06 & 4.46 & 10.15 & 13.77 & 20.61 &35.63 & 44.03 \\
        CLIP-IMG+Q& OKVQA  & 19.08 &11.13 & 34.54 & 50.48 & 65.08 & 80.62 & 88.11  \\
        \midrule
        CLIP-IMG & ReMuQ & 0.28 & 0.11  &0.69& 1.27 & 2.33 & 7.29 & 47.88 \\
        CLIP-Q & ReMuQ & 0.00  & 0.00 & 0.00 & 0.03 & 0.03 & 0.11 & 0.17  \\
        CLIP-IMG+Q& ReMuQ & 0.34 & 0.17 & 0.78 & 1.36 & 2.41 & 7.34 & 47.88  \\
        \bottomrule
    \end{tabular}
    }
    \caption{CLIP performance on two datasets using three approaches to retrieve knowledge. For OKVQA, GS-$112$K corpus is used. P@5 is used for OKVQA and P@1 is used for ReMuQ as shown in the main paper.}
    \label{table:clip_okvqa_result}
\end{table*}

\begin{table*}[t]
    \centering
    \small
    \resizebox{0.8\linewidth}{!}{
    \begin{tabular}{lcccccccc}
        \toprule
        \multirow{2}{*}{\textbf{Model}} & \multirow{2}{*}{\textbf{KB-Size}} & \multicolumn{7}{c}{\textbf{Metric}} \\
        \cmidrule(lr){3-9} 
        ~ & ~  & MRR@$5$  & P@$5$ & R@$5$ & R@$10$ & R@$20$ & R@$50$ & R@$100$  \\
        \midrule
        ReViz+VL-ICT+OKVQA$^{-}$ & GS-$112$K & $47.82$ & $36.50$ & $66.49$ & $77.35$ & $86.23$ & $95.14$ & $95.70$\\
        ReViz+VL-ICT+OKVQA & GS-$112$K & \textbf{54.47 }& \textbf{41.74} & \textbf{73.35} & \textbf{83.17} &\textbf{89.56} & \textbf{94.73} & $96.81$\\
        \midrule
        ReViz+VL-ICT+ReMuQ$^{-}$ & ReMuQ &  50.93& 42.67 & 64.17 & 72.10 & 79.27& 86.81  &90.58 \\
        ReViz+VL-ICT+ReMuQ & ReMuQ &  \textbf{62.11} & \textbf{53.39}& \textbf{76.23} & \textbf{83.32} & \textbf{88.56} & \textbf{93.41} & \textbf{96.12} \\
        \bottomrule
    \end{tabular}
    }
    \caption{Compare the performance of without using hard negative (-) and with hard negative.}
    \label{table:hard_neg}
\end{table*}

\section{Experimental Setup}
All ReViz models consist of a ViLT query encoder and a BERT context encoder, both with 12  transformer blocks with 12 attention heads each. 
For pretraining, we use Adam optimizer with 100 warm-up steps, learning rate at 1e-6, a dropout probability of 0.1, and pre-train the model in 5 epochs. 
For down-stream task fine-tuning, we use Adam optimizer with 10 warm-up steps in 30 epochs. 
Learning rate 1e-6 is applied to fine-tune a pretrained ReViz on the down-stream task, and learning rate 1e-5 is used if fine-tune a vanilla ReViz. 
All models use 64 batch-size in the training on a machine with eight Quadro RTX 8000 GPUs.

\section{Effect of Hard Negative Training}

We show the effectiveness of hard negative training in Table \ref{table:hard_neg}. 
We experiment with both OkVQA and our ReMuQ dataset and the pretrained models on VL-ICT. 
We see that using the hard negative examples to train the model is much better than without this training step.

\section{Additional Visualizations}
\begin{figure*}
\centering
\includegraphics[width=1\textwidth]{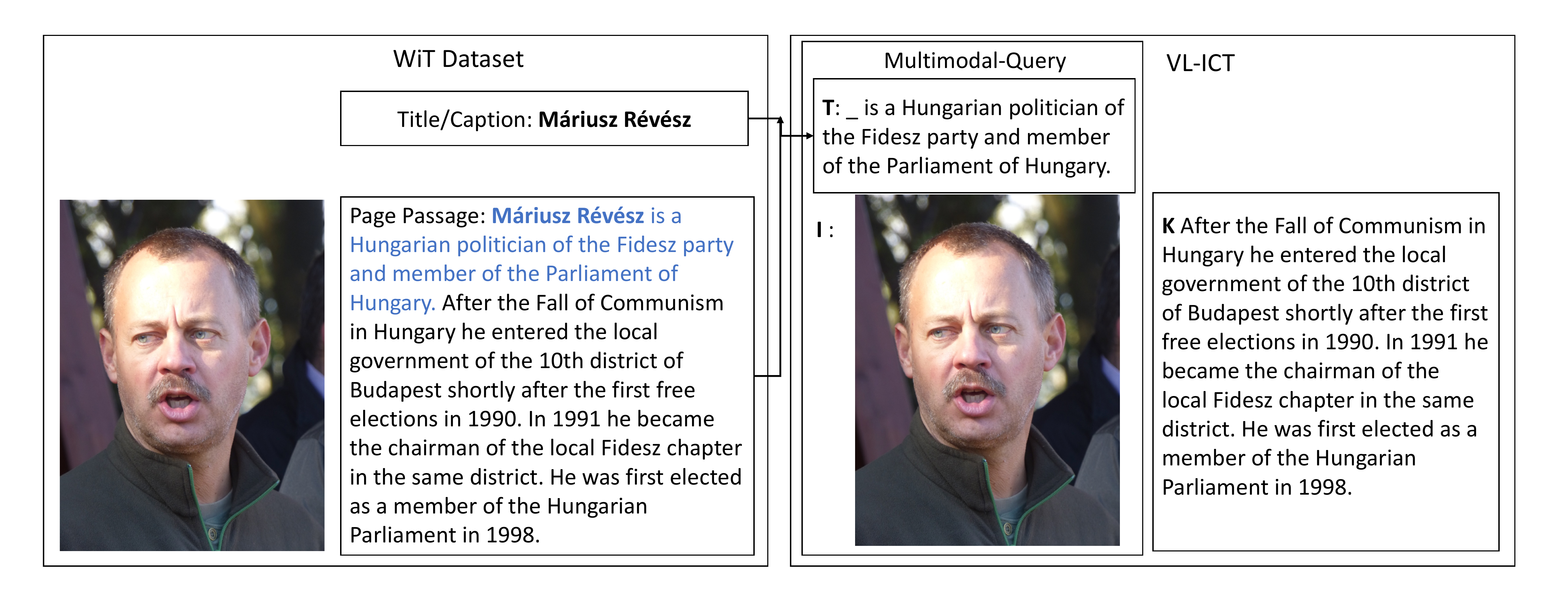}
\includegraphics[width=1\textwidth]{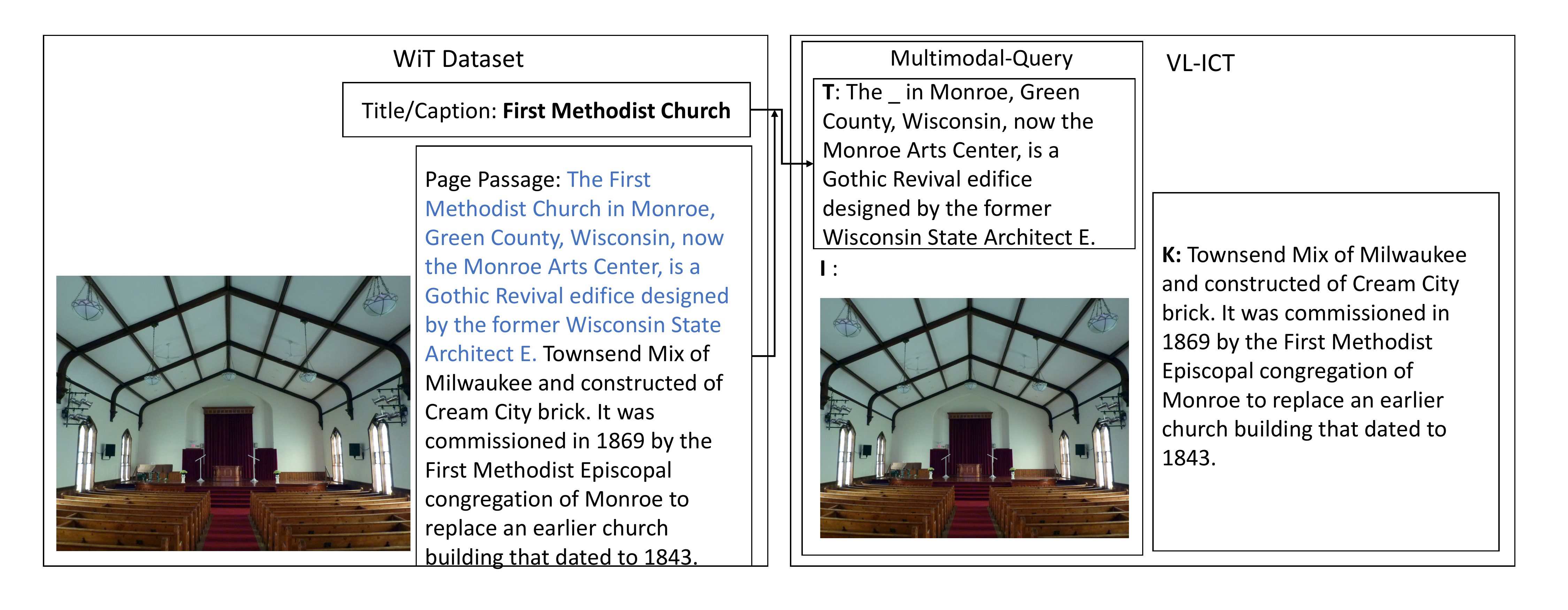}
\caption{More examples of VL-ICT pretraining Task.}
\label{fig:vl_ict_2}
\end{figure*}

\noindent\textbf{Examples of VL-ICT Pretraining Task.}
Figure~\ref{fig:vl_ict_2} presents more examples of VL-ICT pretraining task.

\begin{figure*}[]
\centering
\subfigure{
    \begin{minipage}[t]{0.45\linewidth}
    \centering
    \includegraphics[width=0.95\linewidth, height=4cm]{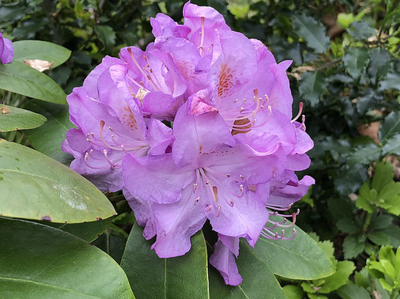}
    \small
    \begin{tabular}{l}
        {\bf Question:} does the flower have petals in a cup shape? \\
        {\bf knowledge:} No, a Minnetonka Rhododendron flower\\ does not have petals in a cup shape. \\
    \end{tabular}
    \end{minipage}
}
\subfigure{
    \begin{minipage}[t]{0.45\linewidth}
    \centering
    \includegraphics[width=0.95\linewidth, height=4cm]{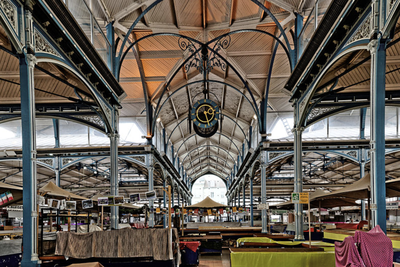}
    \small
    \begin{tabular}{l}
        {\bf Question:} on what is the clock in ceiling of hung on? \\
        {\bf Knowledge:} The clock on the ceiling of Dijon Halles\\ intérieur is hung on a black railing structure. \\
    \end{tabular}
    \end{minipage}
}
\caption{Examples of ReMuQ datasets.}
\label{fig:open_webqa_example}
\end{figure*}

\noindent\textbf{More Examples of ReMuQ Task.}
We present some examples of ReMuQ in Figure \ref{fig:open_webqa_example}, consisting of an image, an input context and the corresponding knowledge. 

\section{Examples of Retrieval Results} 
In Table \ref{tab:okvqa_ret_gs112}, we present some examples of  ReViz+VL-ICT+OKVQA, the best model performing on the GS-112K corpus for OKVQA dataset. 
In Table \ref{tab:okvqa_ret_wiki21}, we present some examples of  ReViz+VL-ICT, the best model performing on the Wiki-21M corpus for OKVQA dataset. 
\begin{table*}[h!]
     \begin{center}
     \resizebox{\linewidth}{!}{
     \begin{tabular}{|c|p{3cm}|p{7cm}|p{3cm}|}
     \hline
      Image & Question & Retrieved Knowledge & Answer \\ \hline
     \raisebox{-\totalheight}{\includegraphics[width=0.3\textwidth, height=40mm]{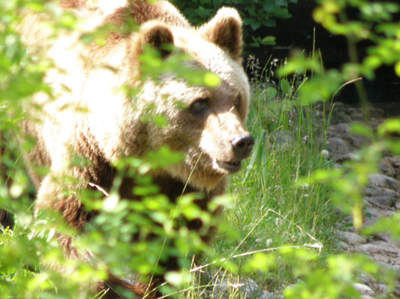}}
      & 
      This bear is what kind of bear?
      & 
      \textbf{brown} bears are found in asia, europe, and north america, giving them the widest ranges of bear species.they also inhabited north africa and the middle east. in north america, \textbf{grizzly} bears previously ranged from alaska down to mexico and as far east as the western shores of hudson bay... 
      &
      grizzly; brown
      \\ \hline
      \raisebox{-\totalheight}{\includegraphics[width=0.3\textwidth, height=40mm]{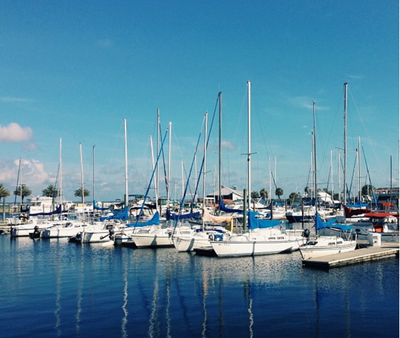}}
      & 
      What do you call the device that keeps boats in place at sea?
      & 
      an \textbf{anchor} is a device, normally made of metal, used to connect a vessel to the bed of a body of water to prevent the craft from drifting due to wind or current.  they have the reputation of not breaking out with tide or wind changes, instead slowly turning in the bottom to align with the force...
      &
      anchor; locator
      \\ \hline
      \raisebox{-\totalheight}{\includegraphics[width=0.3\textwidth, height=40mm]{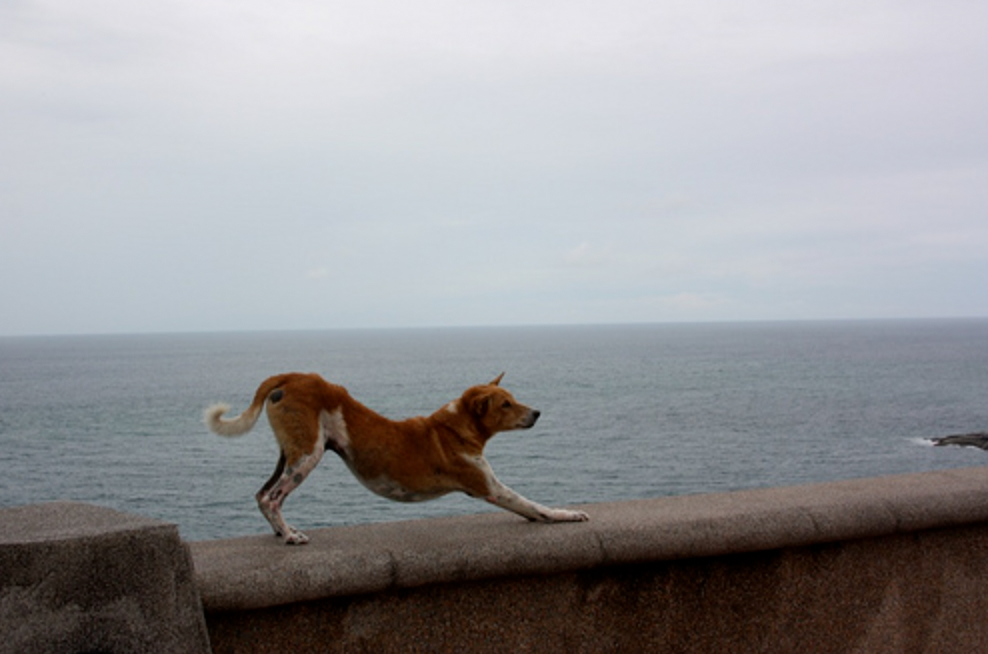}}
      & 
      How long do these animals live?
      & 
      we usually consider 1 year for dog equals to 7 years for you.  similar to dogs, birds' age should be viewed relative to its life span.  different bird species have different life span.
      &
      15 years; fifteen years
      \\ \hline
      \end{tabular}
      }
      \caption{Examples of retrieved knowledge for OKVQA under GS-112K corpus by ReViz+VL-ICT+OKVQA.}
      \label{tab:okvqa_ret_gs112}
\end{center}
\end{table*}

\begin{table*}[h!]
     \begin{center}
     \resizebox{\linewidth}{!}{
     \begin{tabular}{|c|p{3cm}|p{7cm}|p{3cm}|}
     \hline
      Image & Question & Retrieved Knowledge & Answer \\ \hline
     \raisebox{-\totalheight}{\includegraphics[width=0.3\textwidth, height=40mm]{figures/okvqa_1}}
      & 
      This bear is what kind of bear?
      & 
      a 1,600-square mile island. There are currently about 55,000 wild \textbf{grizzly} bears total located throughout North America, most of which reside in Alaska. Only about 1,500 grizzlies are left in the lower 48 states of the US. Of these, about 800 live in Montana...
      &
      grizzly; brown
      \\ \hline
      \raisebox{-\totalheight}{\includegraphics[width=0.3\textwidth, height=40mm]{figures/okvqa_2}}
      & 
      What do you call the device that keeps boats in place at sea?
      & 
      the vessels are not controlled when they arrive in harbor, it depends of many characteristics (old, flag state, cargo, quality of classification society ...). The Memorandum of understanding (MOU) is a group (area) to avoid this restrictive investigation...
      &
      anchor; locator
      \\ \hline
      \raisebox{-\totalheight}{\includegraphics[width=0.3\textwidth, height=40mm]{figures/okvqa_3}}
      & 
      How long do these animals live?
      & 
      in the morning and their dog cries at the window, it exhibits sadness. A growling dog who doesn't like it when someone touches its favorite toy is showing anger. Animals can feel love as well as other basic emotions humans feel. Dogs that grow up with siblings create strong bonds to their sibling.
      &
      15 years; fifteen years
      \\ \hline
      \end{tabular}
      }
      \caption{Examples of retrieved knowledge for OKVQA under Wiki-21M corpus by ReViz+VL-ICT model.}
      \label{tab:okvqa_ret_wiki21}
\end{center}
\end{table*}

\section{CLIP Performance} 

As we mention in the experiment section that CLIP is one of the baselines.  
We compare three methods to retrieve knowledge using CLIP. First one is only using the image, the second one is only by question, and the last one is by both image and question.
In the last method, we firstly use the image embeddings and the knowledge embeddings to obtain the top-100 relevant knowledge, then we use the question embeddings to obtain the top-100 relevant knowledge. Lastly, we obtain the final top-100 knowledge by the sum of the scores given by the image and question embeddings. Table \ref{table:clip_okvqa_result} shows the performance of CLIP using three methods. Using both image and question achieves the best performance.

\end{document}